\documentclass[11pt,a4paper]{article}
\usepackage[hyperref]{eacl2021}
\usepackage{times}
\usepackage{latexsym}
\usepackage{amsmath}
\usepackage{incgraph,tikz}
\usepackage{listings}
\usepackage{color,soul}
\usepackage{todonotes}

\usepackage{graphics,graphicx}

\usepackage{microtype}

\aclfinalcopy

\title{Breaking Writer's Block: \\ Low-cost Fine-tuning of Natural Language Generation Models}

\author{Alexandre Duval \\
        CentraleSupelec \\
\And
  Thomas Lamson \\
  CentraleSupelec \\
  \small{\texttt{\{alexandre.duval, thomas.lamson, gael.de-leseleuc\}@student-cs.fr}} 
\And
  Gaël de Léséleuc de Kérouara \\
  CentraleSupelec \\
 \\\AND
  Matthias Gall\'e \\
  Naver Labs Europe \\
  \small{\texttt{matthias.galle@naverlabs.com}}\\
  }
  
\begin{document}
\maketitle
\begin{abstract}

It is standard procedure these days to solve Information Extraction task by fine-tuning large pre-trained language models. This is not the case for generation task, which relies on a variety of techniques for controlled language generation.

In this paper, we describe a system that fine-tunes a natural language generation model for the problem of solving \textit{Writer's Block}.
The fine-tuning changes the conditioning to also include the right context in addition to the left context, as well as an optional list of entities, the size, the genre and a summary of the paragraph that the human author wishes to generate.

Our proposed fine-tuning obtains excellent results, even with a small number of epochs and a total cost of USD $150$. The system can be accessed as a web-service,\footnote{\url{http://textgen.thomas-lamson.com/}} and all the code is released.\footnote{\url{https://github.com/ThomasLamsonFr/AITextGenerator}} A video showcasing the interface and the model is also available.\footnote{\url{https://www.youtube.com/watch?v=zwezKGrahK0}} 

\end{abstract}

\begin{figure*}
\centering
\includegraphics[width= \textwidth]{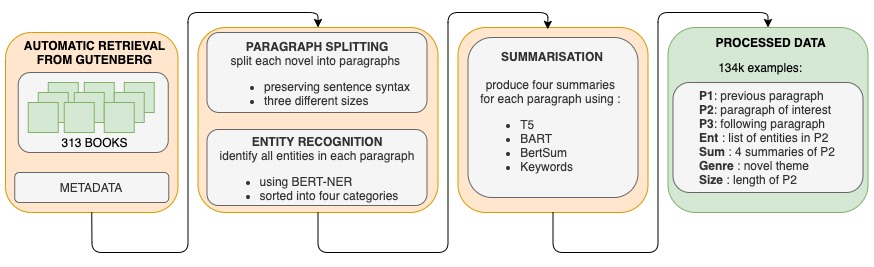}
\caption{Data Preprocessing Pipeline, shows the extracted meta-data that can be used to control the generated text}
\label{fig:data}
\end{figure*}

\section{Introduction}
Thanks to the powerful capacity of large neural networks based on the attention mechanism~\citep{vaswani1706attention}, the current practice in NLP is to start from pre-trained models, which were trained to predict words in context~\cite{devlin2018bert, dai2019transformer} or to perform various other tasks~\citep{raffel2019exploring}.
These pre-trained models are then fine-tuned to solve the task at hand: all top entries of the SuperGLUE benchmark\footnote{\url{https://super.gluebenchmark.com/leaderboard}} for instance follow this trend.

Concerning generation however, the standard methods are very different.
Approaches to \textit{controlled generation} are mostly focused on nudging the model to generate text about a certain topic~\citep{keskar2019ctrl,dathathri2019plug}, or on using distributional models~\citep{khalifa2020distributional}.
Fine-tuning is often dismissed as too expensive as it would require to modify the ensemble of the number of parameters, often measured in the billions.
This is considered impractical, because either too slow, expensive or ecologically not responsible~\citep{strubell2019energy}.
\citet{gpt3} state clearly that ``GPT-3 could also in principle be evaluated in the traditional fine-tuning setting, but we leave this to future work.''
\\

In this paper, we show that it is possible to fine-tune a language model not only to generate text of a certain type, but also to condition it easily on more than a one-sided context. 
In particular, we propose to fine-tune GPT-2 to generate paragraphs based on surrounding (previous and next) sections, a summary of the target content, the entities that should appear, the genre and the desired length.
The resulting model is then used for a web-based system demonstration\footnote{\url{http://textgen.thomas-lamson.com/}} that allows authors to break \textit{Writer's Block}, meaning the ``the condition of being unable to create a piece of written work''.\footnote{\url{https://dictionary.cambridge.org/dictionary/english/writer-s-block}}
Our experiments show that it is possible to obtain excellent results (as measured by a variety of metrics benchmarking the control capacity) with a very limited budget.
The complete training cost of our model, performed on a commercial cloud provider, is around USD $150$.
Our demo introduces the following contribution:

\begin{itemize}
    \item An open-source writing tool that can help creative authors break \textit{Writer's Block}, by proposing novel paragraphs.
    \item A fine-tuned GPT-2 model that respects the context of surrounding paragraphs and allows to control entities, desired output length, the genre as well as the content summary.
    \item Experiments showing that even with a reduced budget, the fine-tuned model diverges from the starting model while generating coherent text.
\end{itemize}

\section{Related Work}

Recent progress in transfer-learning has shown that large pre-trained models are powerful enough to be quickly fine-tuned to solve natural language understanding tasks.

Diverging from this, current approaches to adapt generation models are generally based on picking carefully the \textit{prompt} on which the text is to be generated.
This was popularized by \citet{gpt3} and has since then seen steady growth by different proposals aiming to find good prompts~\citep{schick2020few,schick2020s,gao2020making,li2021prefix}.

A related approach is to adapt the model so that it presents desired biases.
This can be done by training with control tokens~\citep{arivazhagan2019massively,keskar2019ctrl} or by adapting an existing model with additional layers~\citep{kadapters} or sampling techniques~\citep{dathathri2019plug,khalifa2020distributional}.
Those methods allow generating style variations of the same form. 
However, they are less well suited to changes in the conditioned text, such as providing not only a prefix but also a continuation of the text to be generated.
We are particularly interested in conditioning on categorical variables, like {\sc Grover} \citep{zellers2019defending}, but without retraining.
Our experiments show that fine-tuning is surprisingly effective for this.

Several research directions have explored the use of languages models for creative writing and interactive story-telling~\citep{peng-etal-2018-towards, luo-etal-2019-learning}. 
This also includes online tools such as \textit{plot generator}\footnote{https://www.plot-generator.org.uk/story/} or \textit{talk to transformers},\footnote{https://talktotransformer.com/} where the control that can be exerted over the text remains quite rudimentary.
Of special inspiration for this work was \textit{AI Dungeon}.\footnote{https://aidungeon.io/}

\section{Method}
In our approach, the model will be trained to re-generate each book's paragraph (called P2) using the previous and following paragraphs (P1 and P3) as well as information concerning P2: its size, the genre of the book it belongs to, the entities it should include and a summary of its content. 
Instead of training a model from scratch, we leverage a pre-trained GPT-2 117M model and fine-tune it on 313 pre-processed novels. We teach it to predict the next word using the above contextual information as well as already generated words. 

Our approach is separated into three main steps: (i) data preparation (ii) transformation of the data and (iii) fine-tuning:

\subsection{Data}
We emphasise key aspects of the data generation phase, an often overlooked aspect in research projects that however proved essential in our demo.

\paragraph{Novels data} Our paper focuses on text generation for novels and thus requires adequate data. 
We select books from the Gutenberg Project,\footnote{https://www.gutenberg.org/} which we clean and filter based on the associated meta-data.
Only English books corresponding to novels are kept, and the genre (used for fine-tuning later) is defined using a manual mapping from the fine-grained tags provided by Gutenberg.
Due to limited computational resources, we only consider 500 books and ultimately retain 313 after filtering.
We then split the text of each book into paragraphs of different lengths, with a minimum and maximum bound, being careful not to cut a sentence in the middle, nor to separate core parts like chapters or even to split big paragraphs into uneven pieces. 
This step is essential for the later reconstruction within our training phase. 
The size of each paragraph is used to categorise them into \textit{Small} (400-800 characters),  \textit{Medium} (800-1400) or  \textit{Large} (1400-1700).

\paragraph{Entity extraction} Once each book is pre-processed, we detect entities for each paragraph using a pre-trained BERT NER Large model.\footnote{https://github.com/kamalkraj/BERT-NER}
Entities are classified into four categories: persons, locations, organisations and miscellaneous. 
This allows for authors later to control the generation by specifying the entities they wish to incorporate.

\paragraph{Summary} Similarly, in order for authors to be able to guide the generation by giving information on the desired content, we use different summarization models taken from distinct families. This tends to make our model more robust to the possible ways authors could provide this type of information. 
In this sense, we use four different models, covering:
\begin{itemize}
    \item One extractive model (in case authors provide key sentences): BertSum \citep{liu2019fine}.
    \item Two abstractive models (to allow rich re-phrases): BART~\citep{lewis2019bart} and T5~\citep{raffel2019exploring}.
    \item One graph-based non-neural model that extracts keyword phrases:  TextRank~\citep[Kw]{mihalcea2004textrank}
\end{itemize}

\noindent
The full data processing pipeline is shown in Fig.~\ref{fig:data}.

\subsection{Preparation step}

\begin{figure*}[!t]
\centering
\includegraphics[width= \textwidth]{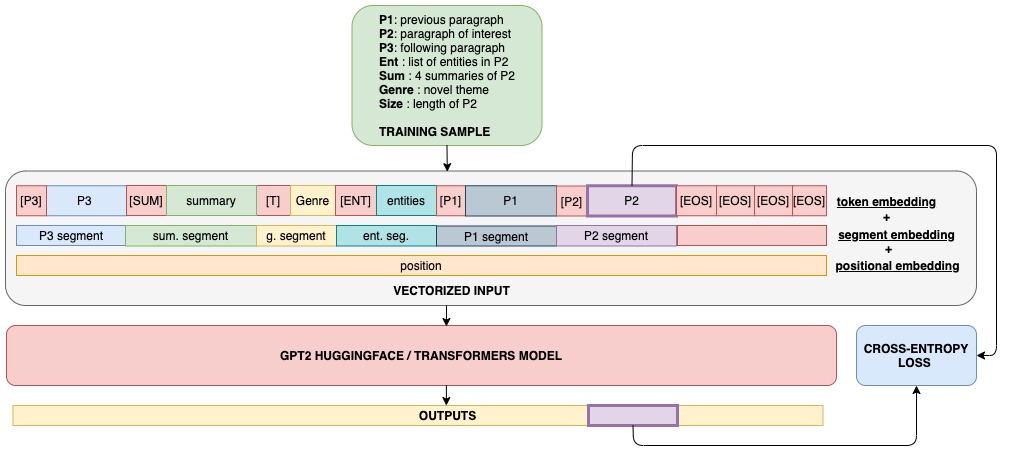}
\caption{Training framework. The loss over the prefix is masked out, and only the cross-entropy loss over P2 is used for fine-tuning.}
\label{training}
\end{figure*}

The resulting documents are split into paragraphs enriched with the related metadata (author, title, language, genre, theme) as well as the four summaries (Bart, T5, BertSum, Kw) and a list of the entities appearing in the text. 
All entities and one summary chosen at random are fed to the GPT-2 model, alongside metadata information (size and genre) and pure text (P1, P2, P3) to help it control and contextualise the generation. \\

The training corpus therefore consists of pairs $(x,y)$ (predict $y$ from prefix $x$), where $y$ is the middle paragraph $P2$ and $x$ is

\begin{verbatim}
   [P3] P3 [Sum] Sum [T] Theme 
   [Ent] Entities [Size] 
   [P1] P1 [P2] 
\end{verbatim}
\noindent where $[P1]$, $[P2]$, $[P3]$, $[Sum]$, $[T]$ and $[Ent]$ indicate the type of input received by the model (special tokens). $[Size]$ is either $[S]$, $[M]$ or $[L]$ and gives information about the paragraph's length. 
Note that the order of the input is not essential. We only put $P1$ at the end so that GPT-2 can continue from there, as it has been trained to do so. \\

The pre-trained model (small GPT-2) has a maximum window size of 1024 tokens.
If $x$ exceeds that length we truncate $P1$ on the left and $P3$ on the right.
As a heuristic we allocate $2/3$ of the remaining space\footnote{once everything except $P1$ and $P3$ has been fed as input} to $P1$ and $1/3$ to $P3$, as we consider $P1$ to be more important than P3.

All the text is segmented using the corresponding pre-trained BPE tokenizer.
Special tokens are created for the separators ($[P1]$, $[P2]$, etc.) and a segment embedding is added on top of the token and position embeddings. It has the same dimension and serves to distinguish the segment each token corresponds to (P1, P2, P3, theme, size, summary and entities). 

\subsection{Fine-tuning}

\begin{figure}[h]
    \centering
    \includegraphics[width = 1 \linewidth]{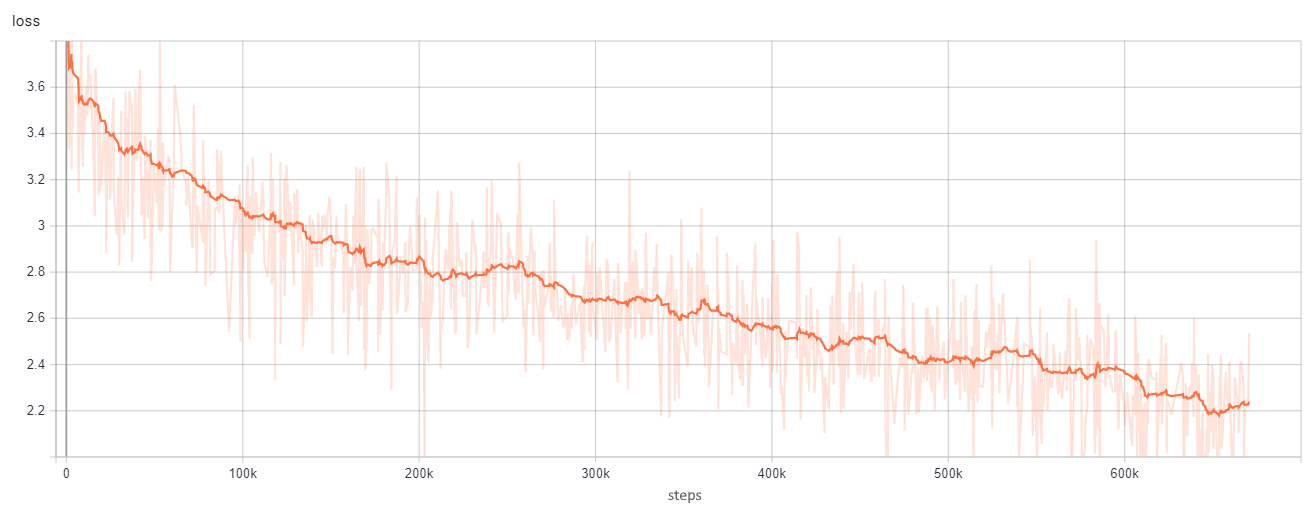}
    \caption{Loss function during training (smoothed).}
    \label{fig:loss}
\end{figure}

We fine-tuned the pre-trained GPT2LMHeadModel (small) from HuggingFace~\citep{huggingface}, using a customised version of the given training script.\footnote{\url{https://huggingface.co/transformers/model\_doc/GPT-2.html}}
$x$ is provided as prefix, and only the cross-entropy error over $y$ (and $P2$) is back-propagated to fine-tune the weights.
The training procedure is shown in Fig.~\ref{training}.
One of the goals of this demo is to show that this type of fine-tuning can be done with limited resources: here we used an AWS's \textit{p3.2xlarge instance} (using one Nvidia Tesla V100 GPU). 
In total, the model received 134k samples for each epoch, and was trained for 10 epochs. 
However, we believe that fewer epochs might be enough to reach good performances although the loss did not converge (Fig.~\ref{fig:loss}).

\section{Web Service Architecture}
The model was enriched with a user interface, and opened to a small targeted public (online community of authors), to gather relevant feedback on both model generation and user-friendliness of the interface.

\begin{figure}
    \centering
    \includegraphics[width = 1 \linewidth]{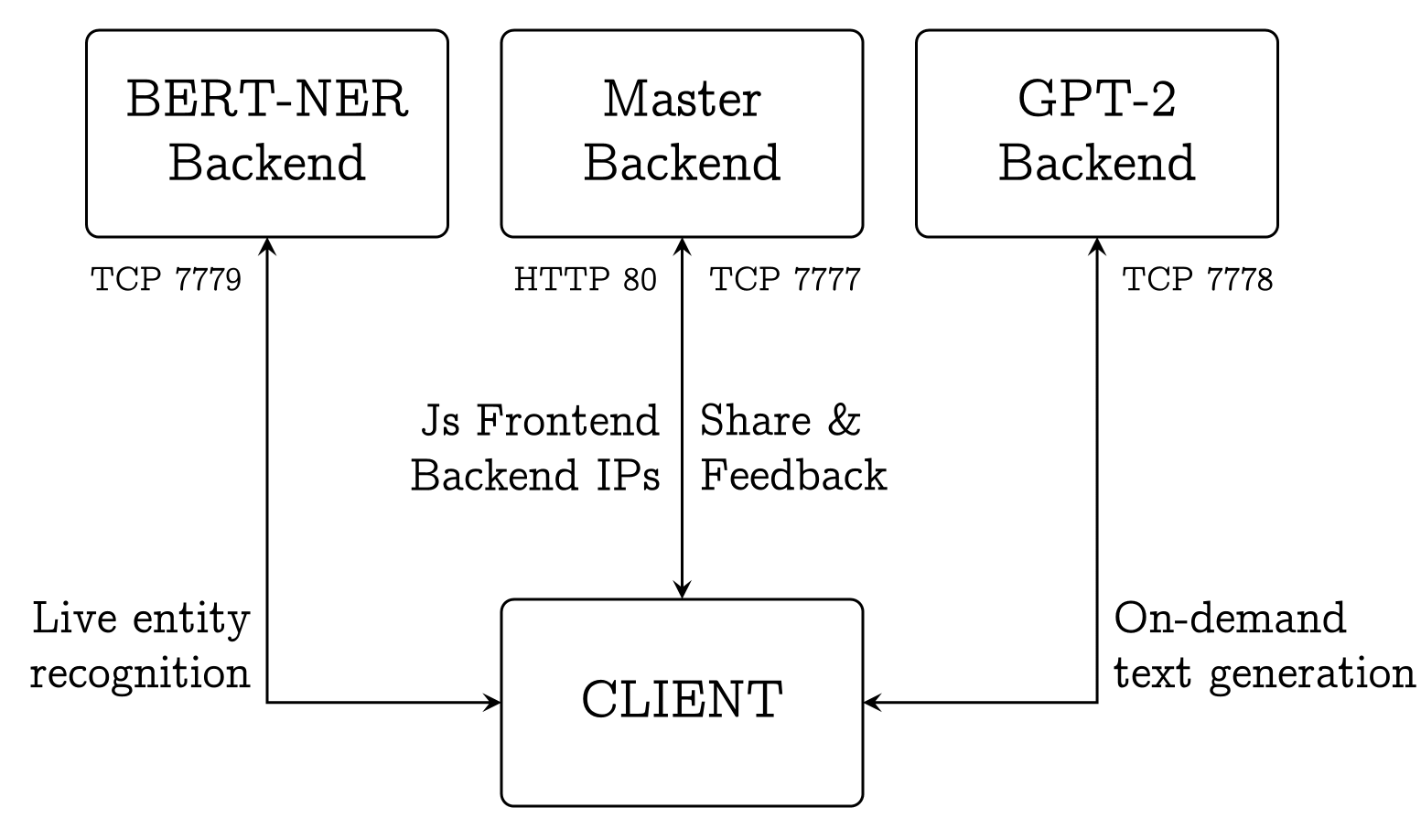}
    \caption{Webservice architecture}
    \label{web}
\end{figure}

To gain in flexibility in the choice of instances, to perform the heavy computations and to allow load balancing on several instances, we uncoupled the \textit{master} instance -- serving the JavaScript front-end and general data -- from the computational instances, performing NER and text generation on demand. 
It is also possible for the client to run the servers locally to avoid delays and server overloads. 
Fig.~\ref{web} shows the general architecture of our service. \\

The interface allows users to write some text in a simple editor. 
Named entities of the four types (characters, locations, organisations and others) are detected on the fly by the NER backend and displayed on the left panel. 
They can be manually edited.

\begin{figure*}[!t]
    \centering
    \includegraphics[width = 250pt]{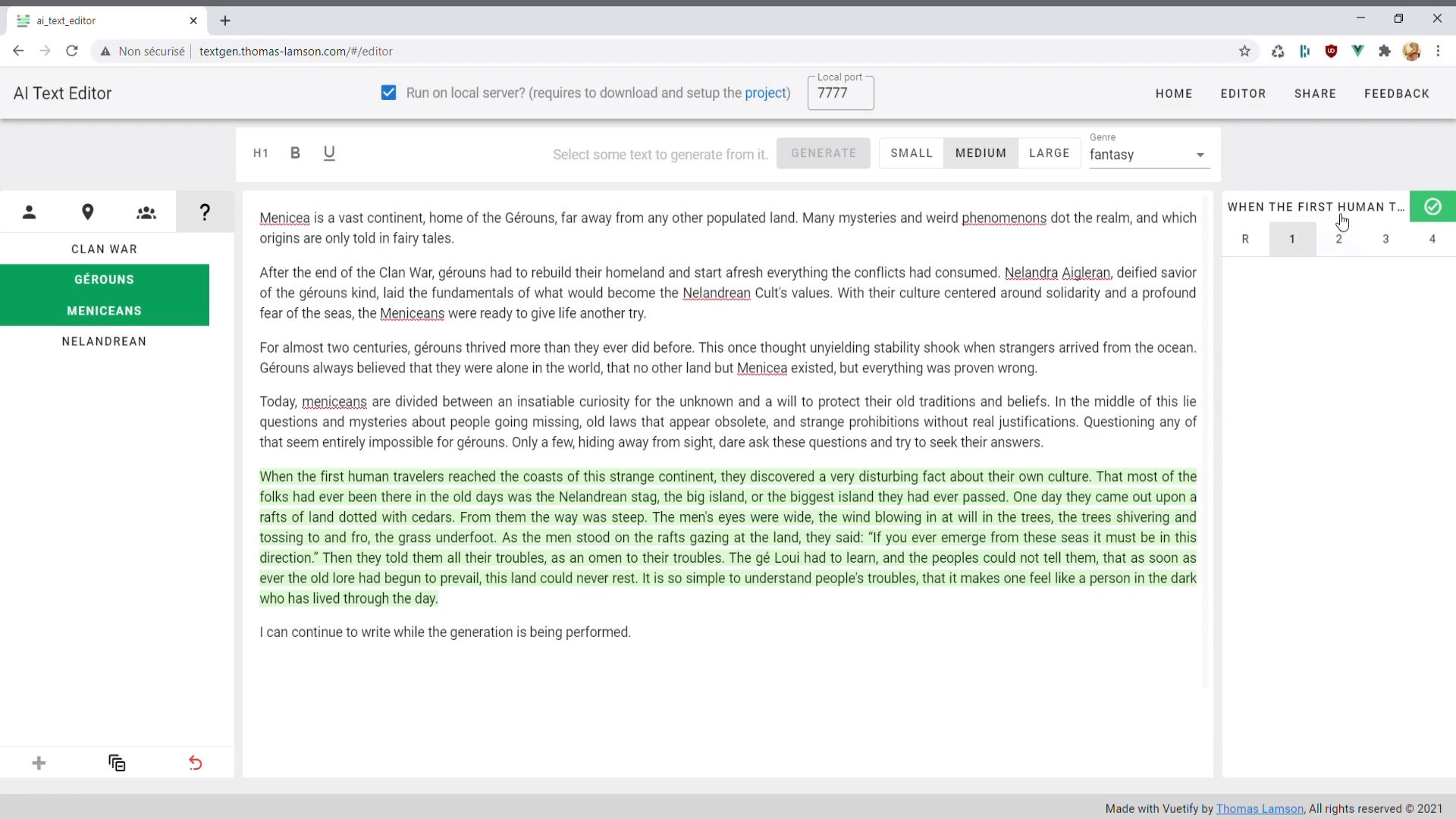}
    \caption{Interface of the text editor, highlightening the generated text in green.}
    \label{fig:interface}
\end{figure*}

Users have the possibility to select several options: length of the desired paragraph, genre of their work and list of entities they want to see appear in the generation. 
They can also highlight a small part of the text that will act as a summary (or a list of keywords).
A snapshot of the interface is shown in Fig.~\ref{fig:interface}.

\section{Generation and Evaluation}

At inference time we provide the prefix $x$ and generate until reaching the end-of-sentence symbol, using Nucleus Sampling~\citep{nucleussampling} with $p=0.9$.

\subsection{Evaluation}

\begin{figure}[h!]
     \centering
     \includegraphics[width=\linewidth]{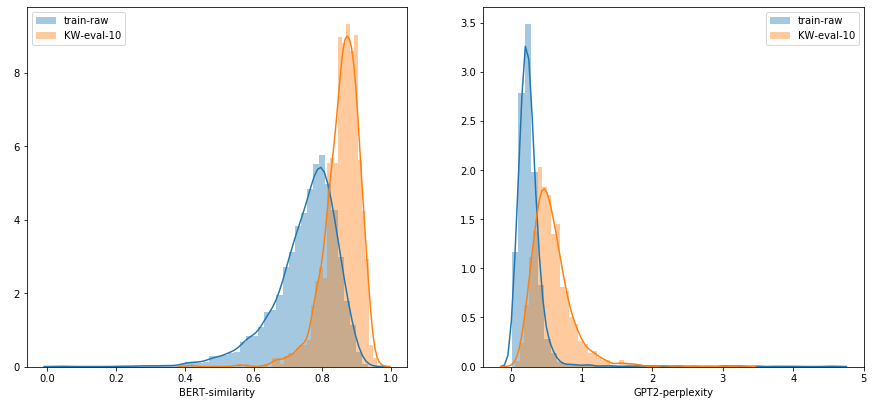}
     \caption{BertSimilarity (left) and Perplexity (right) of the base (not fine-tuned) GPT-2 model and our fine-tuned one.
     Fluency decreases slightly, but the generated text is more similar to the gold middle paragraph.}
     \label{fig:ppx}
 \end{figure}

\begin{figure}[h!]
     \centering
     \includegraphics[width=\linewidth]{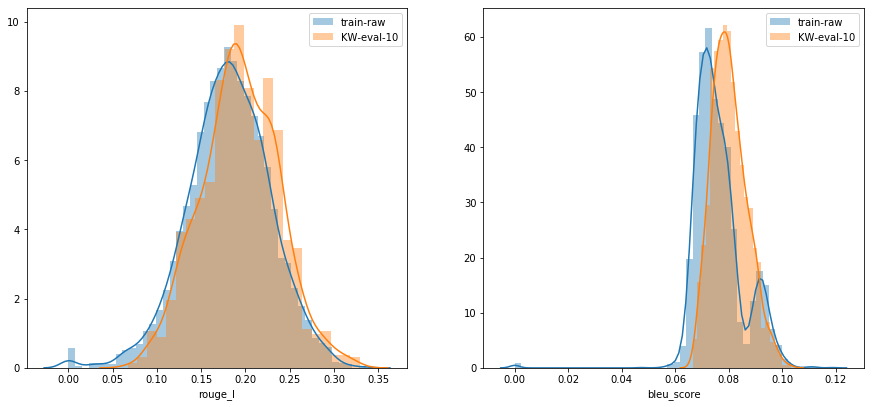}
     \caption{ROUGE (left) and BLEU (right) scores: a small but consistent increase of both metrics.}
     \label{fig:bleu}
 \end{figure}

\begin{figure}[h!]
     \centering
     \includegraphics[width=\linewidth]{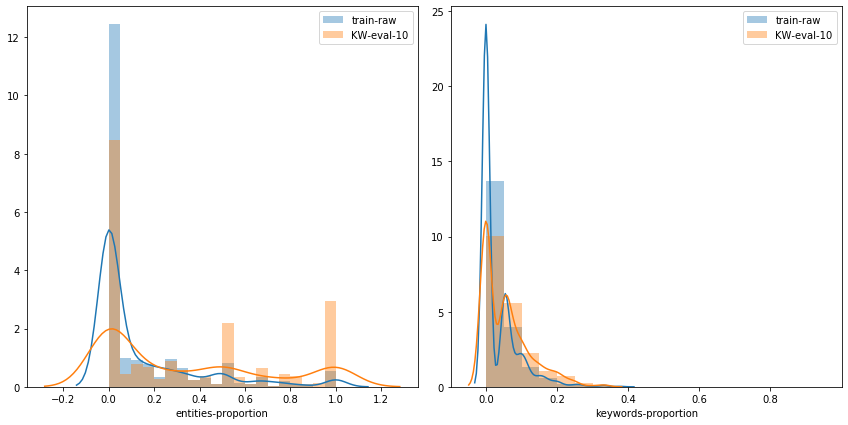}
     \caption{EntitiesCount (left) and KwCount (right). There is a significantly higher proportion of specified entities and keywords appearing in the generated text.}
     \label{fig:control}
 \end{figure}

The final model was evaluated after ten epochs of training, on some unseen novels. 
We focused the evaluation on the degree of control and contextualization, as well as the impact of different types of summaries.
Due to space constraints, we report the results obtained when providing 10 keywords as summaries (extracted with TextRank), but the trend for other summarization techniques is similar.
For the evaluation we focus on
\begin{itemize}
    \item Divergence of the original model, as measured through perplexity of the original GPT-2. (Fig.~\ref{fig:ppx}).
    \item Similarity to the true P2, measured through (i) the similarity of the [CLS] tokens of a pre-trained BERT model~\citep{devlin2018bert} (Fig.~\ref{fig:ppx}) and (ii) BLEU\footnote{we used nltk's version: \url{https://www.nltk.org/_modules/nltk/translate/bleu_score.html}} and ROUGE\footnote{\url{https://pypi.org/project/rouge/}} (Fig.~\ref{fig:bleu}).
    \item Control capabilities, by measuring the number of entities and keywords given as prefix that occur in the resulting text. (Fig.~\ref{fig:control})
\end{itemize}

To evaluate the model, we focus on the distribution of the above metrics across all paragraphs and compare our trained model with a raw GPT-2 model. 

Our experiments show that even with the reduced amount of fine-tuning the model deviates strongly from the base one and is able to learn to produce \textit{middle paragraphs}.

\begin{figure*}[!t]
    \centering
    \includegraphics[width = 0.6 \textwidth]{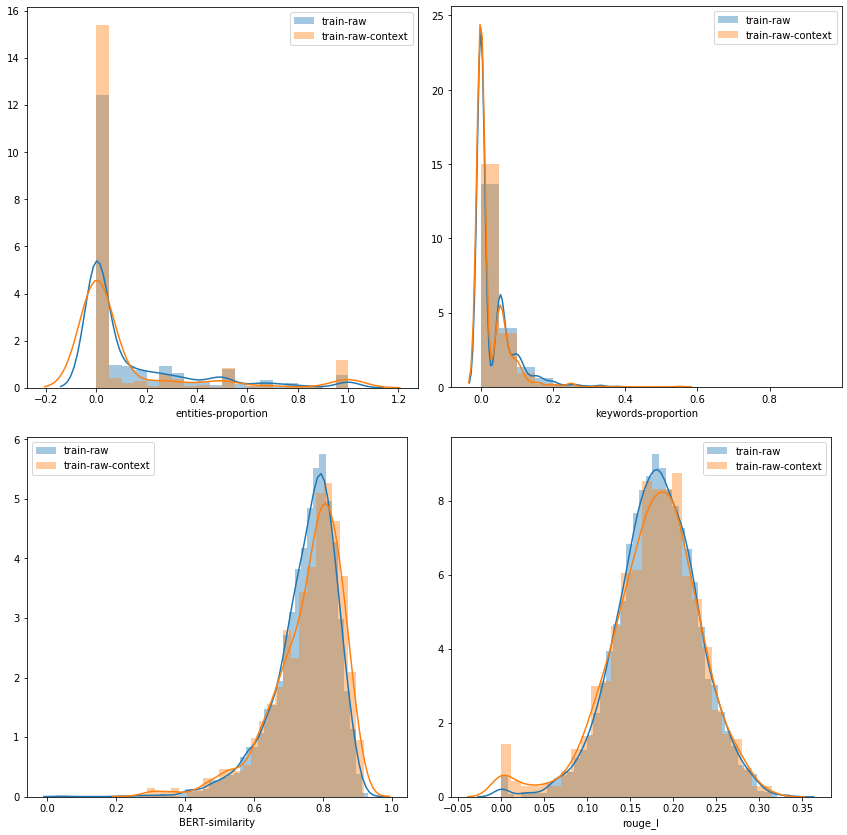}
    \caption{Evaluation metrics on vanilla GPT-2, when providing as prefix only P1 -- \textit{train-raw} -- and the full context (all of $x$) -- \textit{train-raw-context}.}
    
    \label{fig:prompting}
\end{figure*}

Fig~\ref{fig:ppx} shows that our approach leads to a decrease in perplexity (less fluent generation). Nevertheless, this is compensated (of course, those values are not directly comparable) by a better reconstruction of the middle paragraph P2, as shown by the histograms of BERT similarity as well as precision and recall of $n$-gram overlap  Fig.~\ref{fig:bleu}--all significantly shifted to the right. 
Finally, the model clearly learns to control the generated output (Fig.~\ref{fig:control}) with the desired entities occurring most often in the generated text (the shift is weaker with keywords).

As baseline, we also experimented with providing $x$ to the vanilla GPT-2 model.
This allows measuring the added benefit of training with respect to prompting. 
The resulting histograms are shown in Fig.~\ref{fig:prompting}, they reveal that GPT-2 cannot control and contextualise the generation (when taking $x$ as input) if not fine-tuned.

\section{Conclusion}
In this paper, we present an end-to-end pipeline allowing authors to break \textit{Writer's Block}. 
The objective is to allow users -- at any point during the creative writing process -- to generate new paragraphs that are consistent with the rest of the writing, especially previous and following paragraphs. 
The presented tool gives the possibility to select entities (characters, locations, etc.) that have been previously introduced in the novel and that should appear in the target paragraph. 
Similarly, the author can specify the size of the desired text, its content via a small summary or keywords and even the genre of the book. 
In the end, the tool proposes several suggestions that users can choose from and edit. 
The aim is to produce creative outputs that give new ideas to the writers. 

\smallskip

The underlying model is obtained by fine-tuning a GPT-2 model on a carefully designed dataset, obtained through a selection and cleaning of books from the Project Gutenberg library. 
Our experiments show that the generated text is significantly more similar to the gold paragraphs on a variety of metrics and is able to successfully take into consideration the context specified by the user. 

\smallskip

Fine-tuning is often discarded for natural language generation in favour of other cheaper methods, such as prompt engineering or adapter layers.
This work shows a use-case where a pre-trained neural language generation model can be fine-tuned with a reduced economic and ecological cost: the complete training (including preliminary experiments as well as the final mode) was done with a budget of USD $150$.

\bibliography{eacl2021}

\begin{thebibliography}{23}
\expandafter\ifx\csname natexlab\endcsname\relax\def\natexlab#1{#1}\fi

\bibitem[{Arivazhagan et~al.(2019)Arivazhagan, Bapna, Firat, Lepikhin, Johnson,
  Krikun, Chen, Cao, Foster, Cherry et~al.}]{arivazhagan2019massively}
Naveen Arivazhagan, Ankur Bapna, Orhan Firat, Dmitry Lepikhin, Melvin Johnson,
  Maxim Krikun, Mia~Xu Chen, Yuan Cao, George Foster, Colin Cherry, et~al.
  2019.
\newblock Massively multilingual neural machine translation in the wild:
  Findings and challenges.
\newblock \emph{arXiv preprint arXiv:1907.05019}.

\bibitem[{Brown et~al.(2020)Brown, Mann, Ryder, Subbiah, Kaplan, Dhariwal,
  Neelakantan, Shyam, Sastry, Askell et~al.}]{gpt3}
Tom~B Brown, Benjamin Mann, Nick Ryder, Melanie Subbiah, Jared Kaplan, Prafulla
  Dhariwal, Arvind Neelakantan, Pranav Shyam, Girish Sastry, Amanda Askell,
  et~al. 2020.
\newblock Language models are few-shot learners.
\newblock \emph{arXiv preprint arXiv:2005.14165}.

\bibitem[{Dai et~al.(2019)Dai, Yang, Yang, Carbonell, Le, and
  Salakhutdinov}]{dai2019transformer}
Zihang Dai, Zhilin Yang, Yiming Yang, Jaime Carbonell, Quoc~V Le, and Ruslan
  Salakhutdinov. 2019.
\newblock Transformer-xl: Attentive language models beyond a fixed-length
  context.
\newblock \emph{arXiv preprint arXiv:1901.02860}.

\bibitem[{Dathathri et~al.(2019)Dathathri, Madotto, Lan, Hung, Frank, Molino,
  Yosinski, and Liu}]{dathathri2019plug}
Sumanth Dathathri, Andrea Madotto, Janice Lan, Jane Hung, Eric Frank, Piero
  Molino, Jason Yosinski, and Rosanne Liu. 2019.
\newblock Plug and play language models: a simple approach to controlled text
  generation.
\newblock \emph{arXiv preprint arXiv:1912.02164}.

\bibitem[{Devlin et~al.(2018)Devlin, Chang, Lee, and
  Toutanova}]{devlin2018bert}
Jacob Devlin, Ming-Wei Chang, Kenton Lee, and Kristina Toutanova. 2018.
\newblock Bert: Pre-training of deep bidirectional transformers for language
  understanding.
\newblock \emph{arXiv preprint arXiv:1810.04805}.

\bibitem[{Gao et~al.(2020)Gao, Fisch, and Chen}]{gao2020making}
Tianyu Gao, Adam Fisch, and Danqi Chen. 2020.
\newblock \href {http://arxiv.org/abs/2012.15723} {Making pre-trained language
  models better few-shot learners}.

\bibitem[{Holtzman et~al.(2019)Holtzman, Buys, Du, Forbes, and
  Choi}]{nucleussampling}
Ari Holtzman, Jan Buys, Li~Du, Maxwell Forbes, and Yejin Choi. 2019.
\newblock The curious case of neural text degeneration.
\newblock \emph{arXiv preprint arXiv:1904.09751}.

\bibitem[{Keskar et~al.(2019)Keskar, McCann, Varshney, Xiong, and
  Socher}]{keskar2019ctrl}
Nitish~Shirish Keskar, Bryan McCann, Lav~R Varshney, Caiming Xiong, and Richard
  Socher. 2019.
\newblock Ctrl: A conditional transformer language model for controllable
  generation.
\newblock \emph{arXiv preprint arXiv:1909.05858}.

\bibitem[{Khalifa et~al.(2021)Khalifa, Elsahar, and
  Dymetman}]{khalifa2020distributional}
Muhammad Khalifa, Hady Elsahar, and Marc Dymetman. 2021.
\newblock A distributional approach to controlled text generation.
\newblock In \emph{ICLR}.

\bibitem[{Lewis et~al.(2019)Lewis, Liu, Goyal, Ghazvininejad, Mohamed, Levy,
  Stoyanov, and Zettlemoyer}]{lewis2019bart}
Mike Lewis, Yinhan Liu, Naman Goyal, Marjan Ghazvininejad, Abdelrahman Mohamed,
  Omer Levy, Ves Stoyanov, and Luke Zettlemoyer. 2019.
\newblock Bart: Denoising sequence-to-sequence pre-training for natural
  language generation, translation, and comprehension.
\newblock \emph{arXiv preprint arXiv:1910.13461}.

\bibitem[{Li and Liang(2021)}]{li2021prefix}
Xiang~Lisa Li and Percy Liang. 2021.
\newblock Prefix-tuning: Optimizing continuous prompts for generation.
\newblock \emph{arXiv preprint arXiv:2101.00190}.

\bibitem[{Liu(2019)}]{liu2019fine}
Yang Liu. 2019.
\newblock Fine-tune bert for extractive summarization.
\newblock \emph{arXiv preprint arXiv:1903.10318}.

\bibitem[{Luo et~al.(2019)Luo, Dai, Yang, Liu, Chang, Sui, and
  Sun}]{luo-etal-2019-learning}
Fuli Luo, Damai Dai, Pengcheng Yang, Tianyu Liu, Baobao Chang, Zhifang Sui, and
  Xu~Sun. 2019.
\newblock \href {https://doi.org/10.18653/v1/P19-1603} {Learning to control the
  fine-grained sentiment for story ending generation}.
\newblock In \emph{Proceedings of the 57th Annual Meeting of the Association
  for Computational Linguistics}, pages 6020--6026, Florence, Italy.
  Association for Computational Linguistics.

\bibitem[{Mihalcea and Tarau(2004)}]{mihalcea2004textrank}
Rada Mihalcea and Paul Tarau. 2004.
\newblock Textrank: Bringing order into text.
\newblock In \emph{Proceedings of the 2004 conference on empirical methods in
  natural language processing}, pages 404--411.

\bibitem[{Peng et~al.(2018)Peng, Ghazvininejad, May, and
  Knight}]{peng-etal-2018-towards}
Nanyun Peng, Marjan Ghazvininejad, Jonathan May, and Kevin Knight. 2018.
\newblock \href {https://doi.org/10.18653/v1/W18-1505} {Towards controllable
  story generation}.
\newblock In \emph{Proceedings of the First Workshop on Storytelling}, pages
  43--49, New Orleans, Louisiana. Association for Computational Linguistics.

\bibitem[{Raffel et~al.(2019)Raffel, Shazeer, Roberts, Lee, Narang, Matena,
  Zhou, Li, and Liu}]{raffel2019exploring}
Colin Raffel, Noam Shazeer, Adam Roberts, Katherine Lee, Sharan Narang, Michael
  Matena, Yanqi Zhou, Wei Li, and Peter~J Liu. 2019.
\newblock Exploring the limits of transfer learning with a unified text-to-text
  transformer.
\newblock \emph{arXiv preprint arXiv:1910.10683}.

\bibitem[{Schick and Sch{\"u}tze(2020{\natexlab{a}})}]{schick2020few}
Timo Schick and Hinrich Sch{\"u}tze. 2020{\natexlab{a}}.
\newblock Few-shot text generation with pattern-exploiting training.
\newblock \emph{arXiv preprint arXiv:2012.11926}.

\bibitem[{Schick and Sch{\"u}tze(2020{\natexlab{b}})}]{schick2020s}
Timo Schick and Hinrich Sch{\"u}tze. 2020{\natexlab{b}}.
\newblock It's not just size that matters: Small language models are also
  few-shot learners.
\newblock \emph{arXiv preprint arXiv:2009.07118}.

\bibitem[{Strubell et~al.(2019)Strubell, Ganesh, and
  McCallum}]{strubell2019energy}
Emma Strubell, Ananya Ganesh, and Andrew McCallum. 2019.
\newblock Energy and policy considerations for deep learning in nlp.
\newblock \emph{arXiv preprint arXiv:1906.02243}.

\bibitem[{Vaswani et~al.(2017)Vaswani, Shazeer, Parmar, Uszkoreit, Jones,
  Gomez, Kaiser, and Polosukhin}]{vaswani1706attention}
A~Vaswani, N~Shazeer, N~Parmar, J~Uszkoreit, L~Jones, AN~Gomez, L~Kaiser, and
  I~Polosukhin. 2017.
\newblock Attention is all you need. arxiv 2017.
\newblock \emph{arXiv preprint arXiv:1706.03762}.

\bibitem[{Wang et~al.(2020)Wang, Tang, Duan, Wei, Huang, Cao, Jiang, Zhou
  et~al.}]{kadapters}
Ruize Wang, Duyu Tang, Nan Duan, Zhongyu Wei, Xuanjing Huang, Cuihong Cao,
  Daxin Jiang, Ming Zhou, et~al. 2020.
\newblock K-adapter: Infusing knowledge into pre-trained models with adapters.
\newblock \emph{arXiv preprint arXiv:2002.01808}.

\bibitem[{Wolf et~al.(2020)Wolf, Chaumond, Debut, Sanh, Delangue, Moi, Cistac,
  Funtowicz, Davison, Shleifer et~al.}]{huggingface}
Thomas Wolf, Julien Chaumond, Lysandre Debut, Victor Sanh, Clement Delangue,
  Anthony Moi, Pierric Cistac, Morgan Funtowicz, Joe Davison, Sam Shleifer,
  et~al. 2020.
\newblock Transformers: State-of-the-art natural language processing.
\newblock In \emph{Proceedings of the 2020 Conference on Empirical Methods in
  Natural Language Processing: System Demonstrations}, pages 38--45.

\bibitem[{Zellers et~al.(2019)Zellers, Holtzman, Rashkin, Bisk, Farhadi,
  Roesner, and Choi}]{zellers2019defending}
Rowan Zellers, Ari Holtzman, Hannah Rashkin, Yonatan Bisk, Ali Farhadi,
  Franziska Roesner, and Yejin Choi. 2019.
\newblock Defending against neural fake news.
\newblock In \emph{Advances in Neural Information Processing Systems}, pages
  9051--9062.

\end{thebibliography}
\bibliographystyle{acl_natbib}

\newpage

\appendix

\section{Appendix}

\subsection{Interface}

\begin{figure*}[t]
    \centering
    \includegraphics[width = 1 \textwidth]{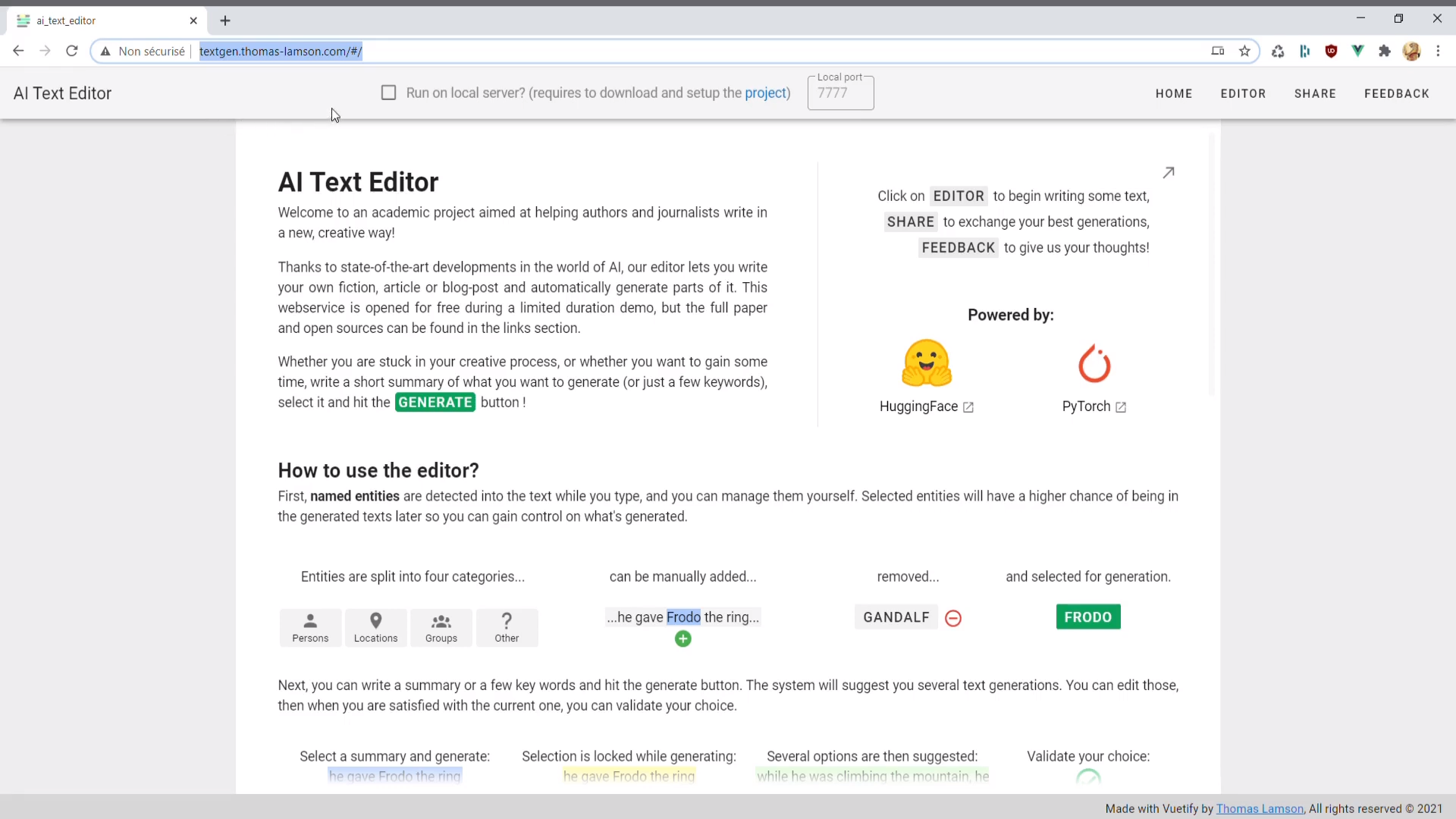}
    \caption{Menu interface}
    \label{menu}
\end{figure*}

Fig.~\ref{menu} shows the landing page of the online demo.

\subsection{Samples of Generated Text}

We show here some samples generated with the demo:

\textbf{Sample 1} \\

\textbf{Keywords:} face - glenure - murder

\textbf{Genre:} biography, history

\textbf{Size:} M

\textbf{Persons:} Glenure, Appin, God, Catriona

\textbf{Organisations:} Appin

\textbf{Locations:} Scotland, Appin, Dean \\

\textbf{P1:} "What is this you say?" she asked. "What are you talking of?" "It is my testimony which may save an innocent life," said I, "and they will not suffer me to bear it. What would you do yourself? You know what this is, whose father lies in danger. Would you desert the poor soul? They have tried all ways with me. They have sought to bribe me; they offered me hills and valleys. \\

\textbf{Generated P2:} But they have not been in the least afraid of me, although they have known me as the \hl{murder}er." "There is something I would have left to have," she said; "if they had taken me from their hands. You shall die." "O my soul!" she cried. "What am I to die at that?" Her father, who had a mother's love, would have killed a slave in the bush; and that was the case. "If you see me here, you'll know the difference between us." He had gone, at least, and she would have heard from him again. There was something he had not understood. "Well," he said, "if that is, then I shall die; if it were not for my sake, I should have been a free man. And if I had done it, I would have died from my father's grave, and taken my money from the country." I had no more idea of his feelings than he had of their manners. On the night appointed, we started for \hl{Scotland}. The men left us a good cabin, and entered, to their great delight. \\

\textbf{Original P2:} And to-day that sleuth-hound told me how I stood, and to what a length he would go to butcher and disgrace me. I am to be brought in a party to the murder; I am to have held Glenure in talk for money and old clothes; I am to be killed and shamed. If this is the way I am to fall, and me scarce a man - if this is the story to be told of me in all Scotland - if you are to believe it too, and my name is to be nothing but a by-word - Catriona, how can I go through with it? The thing's not possible; it's more than a man has in his heart." I poured my words out in a whirl, one upon the other; and when I stopped I found her gazing on me with a startled face. "Glenure! It is the Appin murder," she said softly, but with a very deep surprise. I had turned back to bear her company, and we were now come near the head of the brae above Dean village. At this word I stepped in front of her like one suddenly distracted. "For God's sake!" I cried, "for God's sake, what is this that I have done?" and carried my fists to my temples. \\

\textbf{P3:} "What made me do it? Sure, I am bewitched to say these things!" "In the name of heaven, what ails you now!" she cried. "I gave my honour," I groaned, "I gave my honour and now I have broke it. O, Catriona!" "I am asking you what it is," she said; "was it these things you should not have spoken? And do you think I have no honour, then? or that I am one that would betray a friend?" \\

\newpage

\textbf{Sample 2} \\

\textbf{Genre:} Science-Fiction

\textbf{Size:} L

\textbf{Organisations:} Council

\textbf{Locations:} Council House \\

\textbf{P1:} Along this a disorderly swarm of people marched shouting. They were singing snatches of the song of the revolt, most of them out of tune. Here and there torches flared creating brief hysterical shadows. He asked his way and was twice puzzled by that same thick dialect. His third attempt won an answer he could understand. He was two miles from the wind-vane offices in Westminster, but the way was easy to follow. When at last he did approach the district of the wind-vane offices it seemed to him, from the cheering processions that came marching along the Ways, from the tumult of rejoicing, and finally from the restoration of the lighting of the city, that the overthrow of the Council must already be accomplished. And still no news of his absence came to his ears. The re-illumination of the city came with startling abruptness. Suddenly he stood blinking, all about him men halted dazzled, and the world was incandescent. The light found him already upon the outskirts of the excited crowds that choked the Ways near the wind-vane offices, and the sense of visibility and exposure that came with it turned his colourless intention of joining Ostrog to a keen anxiety. \\

\textbf{Generated P2:} \hl{He could not conceive how he was to get within the unbroken facade of this place} till he reached the street by a great open space of marble pavement to take hold of an afternoon train. For a moment he could not see the crowding. Then another group was seen, then another. \hl{He made his way slowly into the crowding in the central path.} This way he was most eager to know. He had no desire to enter the city in order to get hold of some one else, to whom he could apply as little as possible to information; and so he went at once on the long platform, hoping to reach the \hl{Council House} before he was in time to receive any information that he had received from his brother or his wife. After waiting several minutes for that, the expressman, who had been most unwilling to return to the city, said that he would join the \hl{Council} on the way to meet the new manager, and was even more delighted to have taken possession of his money than he had been when he went to see the \hl{London} cable office and asked about the interview he had held during the afternoon. The cable service had been decided after he had left it. \\

\textbf{Original P2:} For a time he was jostled, obstructed, and endangered by men hoarse and weary with cheering his name, some of them bandaged and bloody in his cause. The frontage of the wind-vane offices was illuminated by some moving picture, but what it was he could not see, because in spite of his strenuous attempts the density of the crowd prevented his approaching it. From the fragments of speech he caught, he judged it conveyed news of the fighting about the Council House. Ignorance and indecision made him slow and ineffective in his movements. For a time he could not conceive how he was to get within the unbroken facade of this place. He made his way slowly into the midst of this mass of people, until he realised that the descending staircase of the central Way led to the interior of the buildings. This gave him a goal, but the crowding in the central path was so dense that it was long before he could reach it. And even then he encountered intricate obstruction, and had an hour of vivid argument first in this guard room and then in that before he could get a note taken to the one man of all men who was most eager to see him. \\

\textbf{P3:} His story was laughed to scorn at one place, and wiser for that, when at last he reached a second stairway he professed simply to have news of extraordinary importance for Ostrog. What it was he would not say. They sent his note reluctantly. For a long time he waited in a little room at the foot of the lift shaft, and thither at last came Lincoln, eager, apologetic, astonished. He stopped in the doorway scrutinising Graham, then rushed forward effusively. "Yes," he cried. "It is you. And you are not dead!" Graham made a brief explanation. "My brother is waiting," explained Lincoln. "He is alone in the wind-vane offices. We feared you had been killed in the theatre. He doubted - and things are very urgent still in spite of what we are telling them there - or he would have come to you." They ascended a lift, passed along a narrow passage, crossed a great hall, empty save for two hurrying messengers, and entered a comparatively little room, whose only furniture was a long settee and a large oval disc of cloudy, shifting grey, hung by cables from the wall. \\

\textbf{BART:} He could not conceive how he was to get within the unbroken facade of this place. He made his way slowly into The crowding in a central path prevented him from reaching it until after an hour's argument with one man who had been most eager for information about what happened at Council House, and then another

\end{document}